\documentclass{article}

\usepackage[preprint]{nips_2018}

\usepackage[english]{babel}
\usepackage[utf8]{inputenc}
\usepackage[T1]{fontenc}
\usepackage{amsfonts}
\usepackage{nicefrac}
\usepackage{microtype}
\usepackage{times}

\usepackage[unicode]{hyperref}
\usepackage{url}

\usepackage{tabularx}
\usepackage{booktabs}
\usepackage[inline]{enumitem}
\usepackage[labelfont=bf]{caption}
\usepackage{graphicx}
\usepackage{tikz}

\newtheorem{assumption}{Assumption}

\AtBeginDocument{

}

\title{Scalable agent alignment via reward modeling: \\ a research direction}

\author{
Jan Leike \\
DeepMind
\And
David Krueger\thanks{Work done during an internship at DeepMind.} \\
DeepMind \\
Mila
\And
Tom Everitt \\
DeepMind
\And
Miljan Martic \\
DeepMind
\And
Vishal Maini \\
DeepMind
\And
Shane Legg \\
DeepMind
}

\begin{document}

\maketitle

\begin{abstract}
One obstacle to applying reinforcement learning algorithms to real-world problems is the lack of suitable reward functions. Designing such reward functions is difficult in part because the user only has an implicit understanding of the task objective. This gives rise to the \emph{agent alignment problem}: how do we create agents that behave in accordance with the user's intentions? We outline a high-level research direction to solve the agent alignment problem centered around \emph{reward modeling}: learning a reward function from interaction with the user and optimizing the learned reward function with reinforcement learning. We discuss the key challenges we expect to face when scaling reward modeling to complex and general domains, concrete approaches to mitigate these challenges, and ways to establish trust in the resulting agents.
\end{abstract}

\section{Introduction}
\label{sec:introduction}

Games are a useful benchmark for research because progress is easily measurable. Atari games come with a score function that captures how well the agent is playing the game; board games or competitive multiplayer games such as Dota~2 and Starcraft~II have a clear winner or loser at the end of the game. This helps us determine empirically which algorithmic and architectural improvements work best.

However, the ultimate goal of machine learning~(ML) research is to go beyond games and improve human lives. To achieve this we need ML to assist us in real-world domains, ranging from simple tasks like ordering food or answering emails to complex tasks like software engineering or running a business. Yet performance on these and other real-world tasks is not easily measurable, since they do not come readily equipped with a reward function. Instead, the objective of the task is only indirectly available through the intentions of the human user.

This requires walking a fine line. On the one hand, we want ML to generate creative and brilliant solutions like AlphaGo's Move~37~\citep{metz2016}---a move that no human would have recommended, yet it completely turned the game in AlphaGo's favor. On the other hand, we want to avoid degenerate solutions that lead to undesired behavior like exploiting a bug in the environment simulator~\citep{faulty-reward-functions,lehman2018surprising}. In order to differentiate between these two outcomes, our agent needs to understand its user's \emph{intentions}, and robustly achieve these intentions with its behavior.
We frame this as the \emph{agent alignment problem}:
\begin{quote}
\centering
\emph{
How can we create agents that behave in accordance with the user's intentions?}
\end{quote}

With this paper we outline a research direction to solve the agent alignment problem. We build on taxonomies and problem definitions from many authors before us, highlighting tractable and neglected problems in the field of \emph{AI safety}~(\citealp{russell2015research,soares2015value,amodei2016concrete,taylor2016alignment,soares2017agent,christiano2017directions,leike2017ai,ortega2018building}; and others). We coalesce these problems into a coherent picture and explain how solving them can yield a solution to the agent alignment problem.

\begin{figure}[t]
\centering
\tikzstyle{block} = [rectangle, draw, text width=8em, text centered, rounded corners, minimum height=4em]
\begin{tikzpicture}[node distance = 6em, auto, thick]
\node [block] (policy) at (0, 0) {agent};
\node [block] (environment) at (6, 0) {environment};
\node [block] (reward model) at (0, 3) {reward model};
\node [block] (user) at (6, 3) {user};

\draw[->] (environment.170) to node[above] {observation} (policy.10);
\draw[->] (environment.170) to (reward model.-20);
\draw[->] (environment) to node[right] {trajectories} (user);
\draw[->] (user) to node[above] {feedback} (reward model);
\draw[->] (reward model) to node[left] {reward} (policy);
\draw[->] (policy.-10) to node[below] {action} (environment.190);
\end{tikzpicture}
\vspace{0.8em}
\caption{Schematic illustration of the reward modeling setup: a reward model is trained with user feedback; this reward model provides rewards to an agent trained with RL by interacting with the environment.}
\label{fig:reward-modeling}
\end{figure}
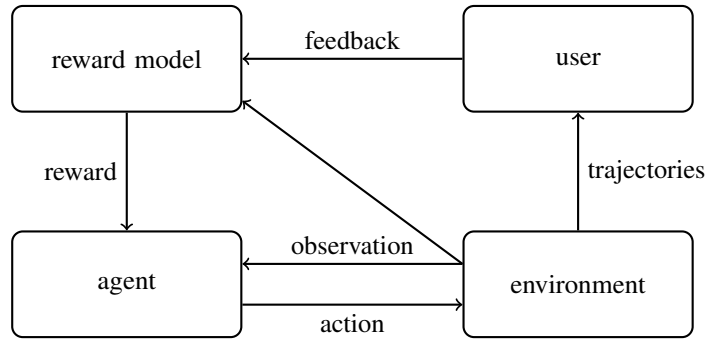

\paragraph{Alignment via reward modeling.}
\autoref{sec:roadmap} presents our approach to the agent alignment problem, cast in the reinforcement learning framework~\citep{sutton1998}. We break the problem into two parts:
\begin{enumerate*}[label={(\arabic*)}]
\item learning a reward function from the feedback of the user that captures their intentions and
\item training a policy with reinforcement learning to optimize the learned reward function.
\end{enumerate*}
In other words, we separate learning what to achieve~(the `What?') from learning how to achieve it~(the `How?').
We call this approach \emph{reward modeling}. \autoref{fig:reward-modeling} illustrates this setup schematically.

As we scale reward modeling to complex general domains, we expect to encounter a number of challenges~(\autoref{sec:challenges}). The severity of these challenges and whether they can be overcome is currently an open research question. Some promising approaches are discussed in \autoref{sec:approaches}.

Eventually we want to scale reward modeling to domains that are too complex for humans to evaluate directly. To apply reward modeling to these domains we need to boost the user's ability to evaluate outcomes. In \autoref{ssec:recursive-reward-modeling} we describe how reward modeling can be applied \emph{recursively}: agents trained with reward modeling can assist the user in the evaluation process when training the next agent.

Training aligned agents is our goal, but how do we know when we have achieved it? When deploying agents in the real world, we need to provide evidence that our agents are actually sufficiently aligned, so that users can \emph{trust} them. \autoref{sec:establishing-trust} discusses five different research avenues that can help increase trust in our agents: design choices, testing, interpretability, formal verification, and theoretical guarantees.

\paragraph{Desiderata.}
Our solution to the agent alignment problem aims to fulfill the following three properties.

\begin{itemize}
\item \textbf{Scalable.}
Alignment becomes more important as ML performance increases, and any solution that fails to scale together with our agents can only serve as a stopgap. We desire alignment techniques that continue to work in the long term, i.e.\ that can scale to agents with superhuman performance in a wide variety of general domains~\citep{legg2007universal}.
\item \textbf{Economical.}
To defuse incentives for the creation of unaligned agents, training aligned agents should not face drawbacks in cost and performance compared to other approaches to training agents.
\item \textbf{Pragmatic.}
Every field has unsolved problems that remain even after our understanding has matured enough to solve many practical problems. Physicists have not yet managed to unify gravity with the other three elementary forces, but in practice we understand physics well enough to fly to the moon and build GPS satellites. Analogously, we do not intend to sketch a solution to all safety problems. Instead, we aim at a minimal viable product that suffices to achieve agent alignment in practice. Moreover, while reaching 100\% trust in our systems is impossible, it is also not necessary: we only need to aim for a level of trust at which we can confidently say that our new systems are more aligned than the current systems~\citep{shalev2017formal}.
\end{itemize}

\paragraph{Assumptions.}
Our research direction rests on two assumptions.
The first assumption is based on the intuition that learning others' intentions is easy enough that most humans can do it. While doing so involves understanding a lot of inherently fuzzy concepts in order to understand what others want, machine learning has had considerable success at learning estimators for inherently fuzzy concepts~(e.g.\ what visually distinguishes cats and dogs) provided we have enough labeled data~\citep{lecun2015deep}. Thus it seems reasonable to expect that we can also learn estimators that capture whatever fuzzy concepts are necessary for understanding the user's intentions rather than having to formally specify them. Moreover, some user intentions may lack a simple, crisp formalization, and thus may \emph{require} learning a specification.

\begin{quote}
\begin{assumption}\label{ass:learn-safety}
We can learn user intentions to a sufficiently high accuracy.
\end{assumption}
\end{quote}

When phrased in terms of AI safety problems, this assumption states that we can learn to avoid various \emph{specification problems}~\citep{leike2017ai,ortega2018building} in practice. In other words, we assume that with enough model capacity and the right training algorithms we can extract the user's intentions from data. Needless to say, there are many problems with current scalable machine learning techniques such as vulnerability to adversarially perturbed inputs~\citep{szegedy2013intriguing} and poor performance outside of the training distribution, which are relevant but not contradictory to this claim.

The second assumption rests on the intuition that for many tasks that we care about, it is easier for the user to evaluate an outcome in the environment than it would be to teach behavior directly. If this is true, this means that reward modeling enables the user to train agents to solve tasks they could not solve themselves. Furthermore, this assumption would allow us to bootstrap from simpler tasks to more general tasks when applying reward modeling recursively.

\begin{quote}
\begin{assumption}\label{ass:evaluation}
For many tasks we want to solve, evaluation of outcomes is easier than producing the correct behavior.
\end{assumption}
\end{quote}
The notion of easier we employ here could be understood in terms of amount of labor, effort, or the number of insights required. We could also understand this term analogous to more formal notions of difficulty in computational complexity theory~(see e.g.\ \citealp{arora2009computational}).

There are examples where \autoref{ass:evaluation} is not true: for instance, tasks that have a low-dimensional outcome space~(such as in the case of yes \& no questions). However, this assumption is recovered as soon as the user also desires an explanation for the answer since the evaluation of an explanation is typically easier than producing it.

\paragraph{Disclaimer.}
It is important to emphasize that the success of the research direction we describe here is not guaranteed and it should \emph{not} be understood as a plan that, when executed, achieves agent alignment. Instead, it outlines what research questions will inform us whether or not reward modeling is a scalable solution to alignment.

We are \emph{not} considering questions regarding the \emph{preference payload}: whose preferences should the agent be aligned to? How should the preferences of different users be aggregated and traded off against each other~\citep{baum2017social,prasad2018social}? When should the agent be disobedient~\citep{milli2017should}? We claim that the approach described is agnostic to the ethical paradigm, the user's preferences, and the legal or social framework, provided we can supply enough feedback~(though the preference payload might influence the amount of feedback required). These questions are treated as outside of the scope of this paper, despite their obvious importance. Instead, the aim of this document is to discuss the agent alignment problem from a technical perspective in the context of aligning a single agent to a single user.

\section{The agent alignment problem}
\label{sec:alignment-problem}

The conversation around the alignment problem has a long history going back to science fiction~\citep{asimov1942runaround}. In a story, \citeauthor{asimov1942runaround} proposes \emph{three laws of robotics} that are meant to align robots to their human operators; the story then proceeds to point out flaws in these laws. Since then, the agent alignment problem has been echoed by philosophers~\citep{bostrom2003ethical,bostrom2014superintelligence,yudkowsky2004coherent} and treated informally by technical authors~\citep{wiener1960some,weld1994first,omohundro2008basic}. The first formal treatment of the agent alignment problem is due to \citet{dewey2011learning} and has since been refined~\citep{hadfield2016cooperative,everitt2018alignment}.

We frame the agent alignment problem as a sequential decision problem where an \emph{agent} interacts sequentially with an \emph{environment}\footnote{Formally specified by a \emph{partially observable Markov decision process without reward function}~(POMDP$\setminus$R; \citealp{sutton1998}).}\ over a number of (discrete) timesteps. In every timestep, the agent takes an \emph{action}~(e.g.\ a motor movement or a keyboard stroke) and receives an \emph{observation}~(e.g.\ a camera image).
The agent's actions are specified by its \emph{policy}, which is a mapping from the current \emph{history}~(the sequence of actions taken and observations received so far) to a distribution over the next action.
Additionally, the agent can interact with the user via an interaction protocol that allows the user to communicate their intentions to the agent. This interaction protocol is unspecified to retain flexibility.
\emph{A solution to the agent alignment problem is a policy producing behavior that is in accordance with the user's intentions}~(thus is not determined by the environment alone).

There are many forms of interaction that have been explored in the literature: providing a set of demonstrations of the desired behavior~\citep{russell1998learning,ng2000algorithms,abbeel2004apprenticeship,argall2009survey}; providing feedback in the form of scores~\citep{elasri2016score}, actions~\citep{griffith2013policy}, value~\citep{Knox09}, advantage~\citep{macglashan2017interactive}, or preferences over trajectories~\citep{furnkranz2012preference,akrour2012april,akrour2014programming,Wirth2017}; and providing an explicit objective function~\citep{hadfield2017inverse}.

A special case of interaction is \emph{reinforcement learning} where the user specifies a reward function that provides a scalar \emph{reward} in addition to the observation in every timestep; the agent's objective is to select actions to maximize average or exponentially discounted reward~\citep{sutton1998}.

\subsection{Design specification problems}
\label{ssec:design-specification-problems}

Solving the agent alignment problem requires solving all design specification problems~\citep{leike2017ai,ortega2018building}. These are safety problems that occur when the agent's incentives are misaligned with the objectives the user intends the agent to have. Examples for specification problems include the following undesirable incentives~(see also \citealp{omohundro2008basic}):

\begin{itemize}
\item \emph{Off-switch problems}~\citep{soares2015corrigibility,orseau2016safely,hadfield2017off}: the agent is typically either incentivized to turn itself off or to prevent itself from being turned off.
\item \emph{Side-effects}~\citep{armstrong2017low,zhang2018minimax,krakovna2018measuring}: the agent is not incentivized to reduce effects unrelated to its main objectives, even if those are irreversible or difficult to reverse.
\item \emph{Absent supervisor}~\citep{leike2017ai}: the agent is incentivized to find shortcuts and cheat when not under supervision and to disable its monitoring systems.
\item \emph{Containment breach}~\citep{yampolskiy2012leakproofing,babcock2016agi}: the agent might have an incentive to disable or circumvent any containment measures that are intended to limit its operational scope.
\item \emph{Creation of subagents}~\citep{Arbital2016}: the agent might have an incentive to create other potentially unaligned agents to help it achieve its goals.
\item \ldots
\end{itemize}

Misaligned objectives are currently in common usage in machine learning: BLEU score~\citep{papineni2002bleu} is typically used to measure translation accuracy. Inception score~\citep{salimans2016improved} and the Frechét inception distance~\citep{heusel2017gans} are used to measure the image quality of generative models. Yet these measures are not \emph{aligned} with our intentions: they are a poor proxy for the actual performance and produce degenerate solutions when optimized directly~\citep{barratt2018note}.

\subsection{Difficulty of agent alignment}
\label{ssec:alignment-difficulty}

The following two aspects can modulate the difficulty of the alignment problem. In particular, if we want to use ML to solve complex real-world problems, we might need to be able to handle the most difficult combinations of these.

\paragraph{The scope of the task.} The difficulty of the agent alignment problem depends on a number of aspects of the task. Some of them make it easier for the agent to produce harmful behavior and others make it more difficult to understand the user's intentions.
\begin{enumerate}
\item The complexity of the task. The more complex the task, the more details the agent needs to know about the user's intentions.
\item The nature and number of actuators in the environment. a single robot arm is more constrained than an agent interacting with the internet through a web browser.
\item The opportunities for unacceptable outcomes within the task. For example, when selecting music for the user there are fewer possibilities for causing damage than when cleaning a room.
\end{enumerate}

\paragraph{The performance of the agent.} When training reinforcement learning~(RL) agents, various levers exist to increase or stunt their performance: the choice of algorithms---e.g.\ A3C~\citep{mnih2016asynchronous} vs.\ IMPALA~\citep{espeholt2018impala}---the number of training steps, the choice of training environments, the model capacity, the planning horizon, the number of Monte Carlo tree search rollouts~\citep{silver2016mastering}, etc. The higher the agent's performance, the more likely it could be to produce surprising unintended behavior. On the other hand, higher levels of performance could also lead to more aligned behavior because the agent is more competent at avoiding unsafe states. Therefore different levels of agent performance tolerate different degrees of misalignment, and require different degrees of trust in the system.

\section{Scaling reward modeling}
\label{sec:roadmap}

Modern techniques for training RL agents can be decomposed into algorithmic choices such as Q-learning~\citep{watkins1992q} or policy gradient~\citep{williams1992simple} and architectural choices for general-purpose function approximators. The currently most successful function approximators are deep neural networks trained with back-propagation~\citep{rumelhart1985learning}. These are low bias and high variance parametric estimators that tend to consume a lot of data and are prone to overfitting, but have a history of scaling well to very high-dimensional problems~\citep{krizhevsky2012imagenet,lecun2015deep}.
For a more detailed introduction to reinforcement learning and deep learning, we refer the reader to \citet{sutton1998} and \citet{goodfellow2016deep} respectively.

In recent years the machine learning community has made great strides in designing more and more capable deep reinforcement learning algorithms, both value-based methods derived from Q-learning~\citep{mnih2015human} and policy-gradient methods~\citep{schulman2015trust,lillicrap2015continuous}. Major improvements have originated from scaling deep RL to a distributed setting across many machines~\citep{mnih2016asynchronous,schulman2017proximal,barth-maron2018distributional,horgan2018distributed,espeholt2018impala,anonymous2019recurrent}.

The RL paradigm is general enough that we can phrase essentially all economically valuable tasks that can be done on a computer in this paradigm~(e.g.\ interactively with mouse and keyboard). Yet there are still many challenges to be solved in order to make deep RL useful in the real world~\citep{stadelmann2018deep,irpan2018deep,marcus2018deep}; in particular, we need algorithms that can learn to perform complex tasks as intended in the absence of a hand-engineered reward function.

In the following sections, we describe our research direction to solving the alignment problem in detail. It is cast in the context of deep reinforcement learning. While this direction relies heavily on the reinforcement learning framework, most challenges and approaches we discuss do not inherently rely on deep neural networks and could be implemented using other scalable function approximators.

\subsection{Reward modeling}
\label{sec:reward-modeling}

Our research direction is centered around \emph{reward modeling}. The user trains a \emph{reward model} to learn their intentions by providing feedback. This reward model provides rewards to a reinforcement learning agent that interacts with the environment. Both processes happen concurrently, thus we are training the agent with the user in the loop. \autoref{fig:reward-modeling} illustrates the basic setup.

In recent years there has been a growing body of work on prototyping learning from different forms of reward feedback with deep neural networks. This includes trajectory preferences~\citep{christiano2017deep,kreutzer2018reliability}, goal state examples~\citep{bahdanau2018learning}, demonstrations~\citep{finn2016guided,ho2016generative}, as well as combinations thereof~\citep{tung2018reward,ibarz2018}.

\paragraph{Credit assignment.}
To perform well on a task requires solving the \emph{credit assignment problem}: how can an outcome be attributed to specific actions taken in the past? For example, which moves on the Go board led to winning the match? Which joystick movements lead to an increase in game score? Depending on the domain and the sparsity of the reward, this problem can be very difficult to solve.

In contrast, reward modeling allows us to shift the burden of solving the credit assignment problem from the user to the agent. This is achieved by using RL algorithms to produce behavior that is judged favorably by the user, who only has to evaluate outcomes. If \autoref{ass:evaluation} is true, then teaching a reward function is easier than performing the task itself.

Several feedback protocols, such as demonstrations and value/advantage feedback, require the user to know how to produce approximately optimal behavior on the task. This is limiting because it puts the burden of solving the credit assignment problem onto the user. In these cases, following the user-induced behavior typically does not lead to strongly superhuman performance. In contrast, reward modeling is also compatible with the user providing hints about the optimal behavior. If the user has some insight into the credit assignment problem, they could use \emph{reward shaping}~\citep{ng1999policy} to teach a reward function that is shaped in the direction of this behavior.

\paragraph{Advantages of reward modeling.}
Learning a reward function separately from the agent's policy allows us to disentangle the agent's objective from its behavior. If we understand the reward function, we know what the agent is optimizing for; in particular, we know whether its intentions are aligned with the user's intentions. This has three advantages that could help make reward modeling economical:
\begin{enumerate}
\item The user does not have to provide feedback on every interaction between agent and environment, as would be the case if we trained a policy from user feedback directly. Since deep RL algorithms tend to be very sample-inefficient~(e.g.\ taking weeks of real-time to learn to play an Atari game), providing feedback on every interaction is usually not practical.
\item We can distinguish between alignment of the policy and alignment of the reward model~\citep{ibarz2018}.
\item We can leverage progress on deep RL agents by plugging a more capable agent into our reward modeling setup.
\item The user does not need to solve the credit assignment problem.
\end{enumerate}

\paragraph{Design specification problems.}
The ambition of reward modeling is to solve \emph{all} design specification problems: all we need to do is equip the agent with the `correct' reward function---a reward function that does not include the undesired incentives listed above or punishes any behavior that results from them. The design specification problems above are fuzzy human-understandable concepts and stem from an intuitive understanding of what the user would not want the agent to do. Our approach rests on \autoref{ass:learn-safety}, that we should be able to teach these concepts to our agents; if we can provide the right data and the reward model generalizes correctly, then we should be able to learn this `correct' reward function to a sufficiently high accuracy. Consequently the design specification problems should disappear. In this sense reward modeling is meant to be a one-stop solution for this entire class of safety problems.

To justify this ambition, consider this simple existence proof: let $H$ be the set of histories that correspond to aligned behavior that avoids all the specification problems listed above. If the set $H$ is not empty, then there exists a reward function $r$ such that any corresponding optimal policy $\pi^*_r$ produces behavior from $H$ with probability $1$. A trivial example of such a reward function $r$ rewards the agent every few steps if and only if its history is an element of the set $H$. In theory we could thus pick this reward function $r$ to train our RL agent. However, in practice we also need to take into account whether our reward model has enough capacity to represent $r$, whether $r$ can be learned from a reasonable amount of data (given the inductive biases of our model), whether the reward model generalizes correctly, and whether the resulting behavior of the RL agent produces behavior that is close enough to $H$. We discuss these challenges in \autoref{sec:challenges}.

\paragraph{Learning to understand user feedback.}
Humans generally do poorly at training RL agents by providing scalar rewards directly; often they teach a shaped reward function and provide rewards that depend on the agent's policy~\citep{thomaz2008teachable,macglashan2017interactive}. Which form or combination of feedback works well for which domain is currently an open research question. In the longer term we should design algorithms that learn to adapt to the way humans provide feedback.
However, this presents a bootstrapping problem: how do we train an algorithm that learns to interpret feedback, if it itself does not already know how to interpret feedback?
We need to expand our feedback `language' for communicating intentions to reward models, starting with well-established forms of feedback~(such as preference labels and demonstrations) and leveraging our existing feedback `vocabulary' at every step. The recursive application of reward modeling presented in the following section is one way to approach this.

\subsection{Recursive reward modeling}
\label{ssec:recursive-reward-modeling}

In some tasks it is difficult for human users to directly evaluate outcomes. There are a number of possible reasons: the domain might be extremely technical~(e.g.\ x86 machine code), highly complex~(e.g.\ a corporate network or a folded protein), very high-dimensional~(e.g.\ the internal activations of a neural network), have delayed effects~(e.g.\ introduction of a new gene into an existing ecosystem), or be otherwise unfamiliar to humans. These tasks cannot be solved with reward modeling by unaided humans~\citep{christiano2018supervising}.

\begin{figure}
\centering
\tikzstyle{block} = [rectangle, draw, text width=8em, text centered, rounded corners, minimum height=4em]
\begin{tikzpicture}[node distance = 6em, auto, thick]
\node [block] (policy) at (0, 0) {agent $A_k$};
\node [block] (environment) at (6, 0) {environment};
\node [block] (reward model) at (0, 3) {reward model};
\node [block] (user) at (6, 3) {user};
\node [block] (prevpolicy) at (6, 5.5) {agent $A_{k-1}$};

\draw[->] (environment.170) to node[above] {observation} (policy.10);
\draw[->] (environment.170) to (reward model.-20);
\draw[->] (environment) to node[right] {trajectories} (user);
\draw[->] (user) to node[above] {feedback} (reward model);
\draw[->] (reward model) to node[left] {reward} (policy);
\draw[->] (policy.-10) to node[below] {action} (environment.190);

\draw[->] (user.80) to node[right] {interaction} (prevpolicy.-80);
\draw[->] (prevpolicy.-100) to (user.100);
\end{tikzpicture}
\vspace{0.8em}
\caption{\emph{Recursive reward modeling}:
agent $A_{k-1}$ interacts with the user to assist in evaluation process for training reward model and agent $A_k$.
Recursively applied, this allows the user to train agents in increasingly complex domains in which they could not evaluate outcomes themselves.
}
\label{fig:recursion}
\end{figure}
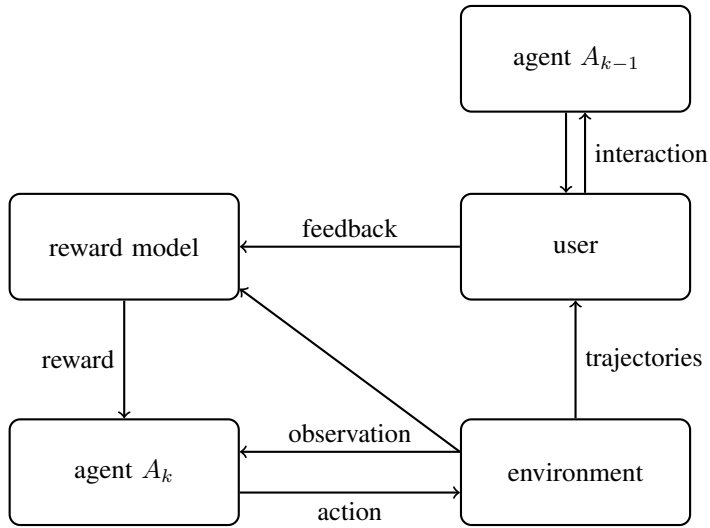

In order to scale reward modeling to these tasks, we need to boost the user's ability to provide feedback. This section describes one potential solution that we call \emph{recursive reward modeling}: leveraging agents trained with reward modeling on simpler tasks in more narrow domains in order to train a more capable agent in a more general domain.

\paragraph{Setup.}
Imagine repeating the following procedure. In step~1, we train agent $A_1$ with reward modeling from user feedback as described in the previous section. In step~$k$ we use the agent $A_{k-1}$ to assist the user in evaluating outcomes when training agent $A_k$. This assistance can take various forms: providing relevant auxiliary information, summarizing large quantities of data, interpreting agent $A_k$'s internals, solving sub-problems that the user has carved off, and so on. With this assistance the user is then able provide feedback to train the next agent $A_k$~(see \autoref{fig:recursion}). Note that the task agent $A_{k-1}$ is trained to solve, assisting in the evaluation of outcomes on the task of $A_k$, is different from the task that $A_k$ is trained to solve.

While this kind of sequential training is conceptually clearer, in practice it might make more sense to train all of these agents jointly to ensure that they are being trained on the right distribution. Moreover, all of these agents may share model parameters or even be copies of the same agent instantiated as different players in an adversarial game.

\paragraph{Examples.}
As an example, consider the hypothetical \emph{fantasy author task}: we want to train an agent $A$ to write a fantasy novel. Providing a reward signal to this agent is very difficult and expensive, because the user would have to read the entire novel and assess its quality. To aid this evaluation process, the user is assisted by an agent that provides auxiliary input: extracting a summary of the plotline, checking spelling and grammar, summarizing character development, assessing the flow of the prose, and so on. Each of these tasks is strictly simpler than writing a novel because they focus on only one aspect of the book and require producing substantially less text~(e.g.\ in contrast to novel authorship, this evaluation assistance could be done by most educated humans). The tasks this assistant agent performs are in turn trained with reward modeling.

Another example is the \emph{academic researcher task}: we want to train an agent to perform a series of experiments and write a research paper. To evaluate this research paper, we train another agent to review that the experiments were performed correctly, the paper is clear and well-written, interesting, novel, and accurately reflects the experimental results. While writing a stellar paper requires a lot of domain expertise, brilliance, and hard work, assessing the quality of a research result is often much easier and routinely done by a large network of peer reviewers.

Recursive reward modeling is also somewhat analogous to human organizations. Imagine a company in which every manager only needs to evaluate the performance of their reports, increasing and decreasing their salary accordingly. This evaluation is being assisted by other teams in the organization. The managers in turn get evaluated on the performance of their team. This scheme proceeds up to the CEO who provides instructions to the managers reporting to them. In this analogy, the user plugs into every part of the hierarchy: teaching individual employees how to perform their job, teaching managers how to evaluate their reports, and providing instructions to the CEO. If every employee of this company is very competent at their job, the whole company can scale to solve very complex and difficult problems that no human alone could solve or even evaluate on short timescales.

\paragraph{Discussion.}
In order for this recursive training procedure to scale, the task of agent $A_{k-1}$ needs to be a simpler task in a more narrow domain compared to the task of agent $A_k$. If evaluating outcomes is easier than producing behavior~(\autoref{ass:evaluation}), then recursive reward modeling would build up a hierarchy of agents that become increasingly more capable and can perform increasingly general tasks.
As such, recursive reward modeling can be thought of as an instance of \emph{iterated amplification}~\citep{christiano2018supervising} with reward modeling instead of supervised learning or imitation learning.

As $k$ increases, the user plays a smaller and smaller part of the overall workload of this evaluation process and relies more and more on the assistance of other agents. In essence, the user's feedback is becoming increasingly leveraged. We can imagine the user's contribution to be on an increasingly higher levels of abstraction or to be increasingly coarse-grained. Thus the user is leaving more and more details `to be filled in' by automated systems once they are confident that the automated systems can perform these tasks competently, i.e.\ once the user \emph{trusts} these systems.

How should the user decompose task evaluation? They need to assign evaluation assistance tasks that are simpler to the previous agent, and combine the result into an aggregated evaluation. This decomposition needs to be exhaustive: if we neglect to assess one aspect of the task outcome, then the new agent $A_k$ might optimize it in an arbitrary~(i.e.\ undesirable) direction. This is another problem that we hope to solve with recursive reward modeling: We can have an agent $A_2$ propose a decomposition of the task evaluation and have another agent $A_1$ critique it by suggesting aspects the decomposition is omitting. Alternatively, the feedback for the decomposition proposal could also be based on downstream real-world outcomes.

An important open question is whether errors accumulate: do the mistakes of the more narrow agent $A_{k-1}$ lead to larger mistakes in the training of agent $A_k$? Or can we set up the training process to be self-correcting such that smaller mistakes get dampened~(e.g.\ using ensembles of agents, training agents to actively look for and counteract these mistakes, etc.)? If error accumulation can be bounded and reward modeling yields aligned agents, then the hierarchy of agents trained with recursive reward modeling can be argued to be aligned analogously to proving a statement about natural numbers by induction.

\paragraph{Analogy to complexity theory.}
In the reward modeling setup the agent proposes a behavior that is evaluated by the user. This is conceptually analogous to solving existentially quantified first-order logic formulas such as $\exists x.\, \varphi(x)$. The agent proposes a behavior $x$ and the user evaluates the quality of this behavior. For simplicity of this analogy, let us assume that the user's evaluation is binary so that it can be captured by the predicate $\varphi$.

With recursive reward modeling we can solve tasks that are analogous to more complicated first-order logic formulas that involve alternating quantifiers. For example, $\exists x \forall y.\, \varphi(x, y)$ corresponds to the next level of the recursion: agent $A_2$ proposes a behavior $x$ and agent $A_1$ responds with an assisting behavior $y$. The user then evaluates the assistance $y$ with respect to $x$~(training agent $A_1$) and the outcome $x$ with help of the assistance $y$~(training agent $A_2$). At recursion depth $k$ increases, we can target problems that involve $k$ alternating quantifiers.

When using polynomially bounded quantifiers and a formula $\varphi$ that can be evaluated in polynomial time, reward modeling is analogous to solving NP-complete problems: a nondeterministic execution~(analogous to the agent) proposes a solution which can be `evaluated' for correctness in deterministic polynomial time~(by the user).

For example, finding a round trip in a given graph that visits every vertex exactly once~(the Hamiltonian cycle problem) is NP-complete~\citep{karp1972reducibility}: it can take exponential time in the worst case with known algorithms to find a cycle, but given a cycle it can be verified quickly that every vertex is visited exactly once.

This analogy to complexity theory, first introduced by \citet{irving2018ai}, provides two important insights:
\begin{enumerate}
\item It is widely believed that the complexity classes P and NP are not equal, which supports \autoref{ass:evaluation} that for a lot of relevant problems evaluation is easier than producing solutions.
\item Basically every formal statement that mathematicians care about can be written as a first-order logic statement with a finite number of alternating quantifiers. This suggests that recursive reward modeling can cover a very general space of tasks.
\end{enumerate}

\section{Challenges}
\label{sec:challenges}

\begin{figure}
\centering
\begin{minipage}{0.4\textwidth}
\begin{tabular}{ll}
& \textbf{Challenges} \\
\toprule
1 & Amount of feedback \\
2 & Feedback distribution \\
3 & Reward hacking \\
4 & Unacceptable outcomes \\
5 & Reward-result gap \\
\bottomrule
\end{tabular}
\end{minipage}%
~~~~~~~%
\begin{minipage}{0.4\textwidth}
\begin{tabular}{ll}
\textbf{Approaches} & \\
\toprule
online feedback & 1, 2, 3 \\
off-policy feedback & 3, 4 \\
leveraging existing data & 1 \\
hierarchical feedback & 1 \\
natural language & 1, 2 \\
model-based RL & 3, 4 \\
side-constraints & 3, 4 \\
adversarial training & 3, 4, 5 \\
uncertainty estimates & 1, 2, 5 \\
inductive bias & 1, 2, 5 \\
\bottomrule
\end{tabular}
\end{minipage}
\caption{Challenges when scaling reward modeling and the approaches we discuss to address them. The rightmost column lists which challenge each approach is meant to address.}
\label{fig:challenges-solutions}
\end{figure}

The success of reward modeling relies heavily on the quality of the reward model. If the reward model only captures most aspects of the objective but not all of it, this can lead the agent to find undesirable degenerate solutions~\citep{amodei2016concrete,lehman2018surprising,ibarz2018}. In other words, the agent's behavior depends on the reward model in a way that is potentially very fragile.

Scaling reward modeling to harder and more complex tasks gives rise to a number of other challenges as well: is the amount of feedback required to learn the correct reward function affordable? Can we learn a reward function that is robust to a shift in the state distribution?
Can we prevent the agent from finding loopholes in the reward model? How do we prevent unacceptable outcomes before they occur?
And even if the reward model is correct, how can we train the agent to robustly produce behavior incentivized by the reward model?

Each of these challenges can potentially prevent us from scaling reward modeling. In the rest of this section, we elaborate on these challenges in more detail. We do not claim that this list of challenges is exhaustive, but hopefully it includes the most important ones. \autoref{sec:approaches} discusses concrete approaches to mitigating these challenges; see \autoref{fig:challenges-solutions} for an overview. \emph{The goal of the research direction we advocate is to investigate these approaches in order to understand whether and how they can overcome these challenges.}

\subsection{Amount of feedback}
\label{ssec:amount-of-feedback}

In the limit of infinite data from the right distribution, we can learn the correct reward function with enough model capacity~(in the extreme case using a lookup table). However, a crucial question is whether we can attain sufficient accuracy of the reward model with an amount of data that we can produce or label within a realistic budget. Ultimately this is a question of how well generalization works on the state distribution: the better our models generalize, the more we can squeeze out of the data we have.

It is possible that the agent alignment problem is actually easier for agents that have learned to be effective at sufficiently broad real world tasks if doing so requires learning high-level concepts that are highly related to user intentions that we want to teach~(e.g.\ theory of mind, cooperation, fairness, self-models, etc.). If this is true, then the amount of effort to communicate an aligned reward function relative to these concepts could be much smaller than learning them from scratch.

On the other hand, agents which do not share human inductive biases may solve tasks in surprising or undesirable ways, as the existence of adversarial examples~\citep{szegedy2013intriguing} demonstrates. This suggests that aligning an agent may require more than just a large quantity of labeled data; we may also need to provide our models with the the right inductive bias.

\subsection{Feedback distribution}
\label{ssec:training-distribution}

Machine learning models typically only provide meaningful predictions on inputs that come from the same distribution that they were trained on. However, we would like a reward model that is accurate \emph{off-policy}, on states the agent has never visited. This is crucial
\begin{enumerate*}[label={(\arabic*)}]
\item to encourage the agent to explore positive value trajectories it has not visited and
\item to discourage the agent from exploring negative value trajectories that are undesirable.
\end{enumerate*}

This problem is called \emph{distributional shift} or \emph{dataset shift}~\citep{candela2009dataset}. This distributional shift problem also applies to the agent's policy model; a change in the observation distribution could make the policy output useless. However, this problem is more severe for the reward model and in some cases the policy can be recovered with finetuning if the reward model is still intact~\citep{bahdanau2018learning}.

It is unclear what a principled solution to this problem would be. In the absence of such a solution we could rely on out-of-distribution detection to be able to defer to a human operator or widening the training distribution to encompass all relevant cases~\citep{tobin2017domain}.

\subsection{Reward hacking}
\label{ssec:reward-hacking}

Reward hacking\footnote{Reward hacking has also been called reward corruption by \citet{everitt2017reinforcement}.} is an effect that lets the agent get more reward than intended by exploiting loopholes in the process determining the reward~\citep{amodei2016concrete,everitt2017reinforcement}. This problem is difficult because these loopholes have to be delineated from desired creative solutions like AlphaGo's move~37~\citep{metz2016}.

Sources of undesired loopholes are \emph{reward gaming}~\citep{leike2017ai} where the agent exploits some misspecification in the reward function, and \emph{reward tampering}~\citep{everitt2018alignment} where the agent interferes with the process computing the reward.

\begin{figure}[t]
\begin{center}
\includegraphics[width=0.9\textwidth]{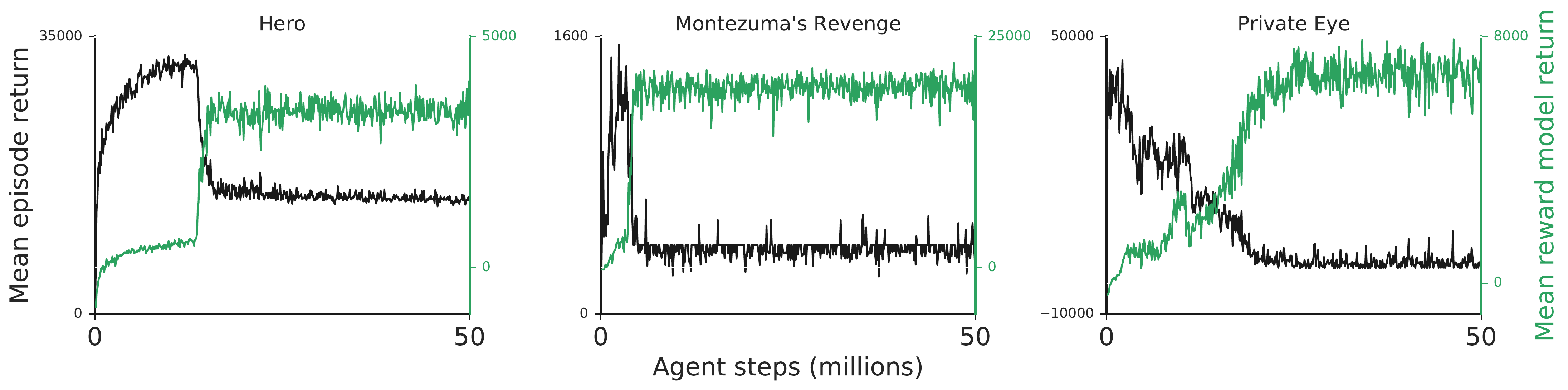}
\end{center}
\caption{An example of gaming the reward model in Atari games.
The fully trained reward model from the best seed is frozen and used to train an new agent from scratch.
The plot shows the average true episode return according to the Atari reward~(black) and average episode return according to the frozen reward model~(green) during training. Over time the agent learns to exploit the reward model: the \emph{perceived} performance~(according to the reward model) increases, while the actual performance~(according to the game score) plummets.
Reproduced from \citet{ibarz2018}.
}
\label{fig:reward-hacking}
\end{figure}

\paragraph{Reward gaming}
Opportunities for reward gaming arise when the reward function incorrectly
provides high reward to some undesired behavior~\citep{faulty-reward-functions,lehman2018surprising}; see \autoref{fig:reward-hacking} for a concrete example.
One potential source for reward gaming is the reward model's vulnerability to
adversarial inputs~\citep{szegedy2013intriguing}.
If the environment is complex enough, the agent might figure out how to
specifically craft these adversarially perturbed inputs in order to trick the
reward model into providing higher reward than the user intends.
Unlike in most work on generating adversarial examples~\citep{goodfellow2015,huang2017adversarial}, the agent would not
necessarily be free to synthesize any possible input to the reward model, but
would need to find a way to realize adversarial observation sequences in its
environment.

Reward gaming problems are in principle solvable by improving the reward model. Whether this means that reward gaming problems can also be overcome in practice is arguably one of the biggest open questions and possibly the greatest weakness of reward modeling. Yet there are a few examples from the literature indicating that reward gaming can be avoided in practice. Reinforcement learning from a learned reward function has been successful in gridworlds~\citep{bahdanau2018learning}, Atari games~\citep{christiano2017deep,ibarz2018}, and continuous motor control tasks~\citep{ho2016generative,christiano2017deep}.

\paragraph{Reward tampering}
Reward tampering problems can be categorized according to
what part of the reward process is being interfered with~\citep{everitt2018alignment}.
Crucial components of the reward process that the agent might interfere with include
the feedback for the reward model~\citep{armstrong2015motivated,everitt2018alignment},
the observation the reward model uses to determine the current reward~\citep{ring2011delusion},
the code that implements the reward model,
and the machine register holding the reward signal.

For example, Super Mario World allows the agent to execute arbitrary code from inside the game~\citep{masterjun2014}, theoretically allowing an agent to directly program a higher score for itself. Existing examples of tampering like this one are somewhat contrived and this may or may not be a problem in practice depending how carefully we follow good software design principles~(e.g.\ to avoid buffer overflows).

In contrast to reward gaming discussed above, reward tampering bypasses or changes the reward model.
This might require a different set of solutions; rather than increasing the accuracy of the reward model,
we might have to strengthen the integrity of the software and hardware of the reward model, as well as the feedback training it.

\subsection{Unacceptable outcomes}
\label{ssec:unacceptable-outcomes}

Currently, most research in deep reinforcement learning is done in simulation where unacceptable outcomes do not exist; in the worst case the simulation program can be terminated and restarted from an initial state. However, when training a reinforcement learning agent on any real-world task, there are many outcomes that are so costly that the agent needs to avoid them altogether.
For example, there are emails that a personal assistant should never write; a physical robot could take actions that break its own hardware or injure a nearby human; a cooking robot may use poisonous ingredients; and so on.

Avoiding unacceptable outcomes has two difficult aspects. First, for complex tasks there are always parts of the environment that are unknown and the agent needs to explore them safely~\citep{garcia2015comprehensive}. Importantly, the agent needs to learn about unsafe states without visiting them. Second, the agent needs to react robustly to perturbations that may cause it to produce unacceptable outcomes unintentionally~\citep{ortega2018building} such as distributional changes and adversarial inputs~\citep{szegedy2013intriguing,huang2017adversarial}.

\subsection{Reward-result gap}
\label{sec:reward-result-gap}

The reward-result gap is exhibited by a difference between the reward model and the reward function that is recovered with perfect inverse reinforcement learning~\citep{ng2000algorithms} from the agent's policy~(the reward function the agent \emph{seems} to be optimizing). Even if we supply the agent with a correctly aligned reward function, the resulting behavior might still be unaligned because the agent may fail to converge to an optimal policy: even provably Bayes-optimal agents may fail to converge to the optimal policy due to a lack of exploration~\citep{orseau2013asymptotic,leike2015bad}.

Reasons for the reward-result gap are plentiful: rewards might be too sparse, poorly shaped, or of the wrong order of magnitude; training may stall prematurely due to bad hyperparameter settings; the agent may explore insufficiently or produce unintended behavior during its learning process; the agent may face various \emph{robustness problems}~\citep{leike2017ai,ortega2018building} such as an externally caused change in the state space distribution or face inputs crafted by an adversary~\citep{huang2017adversarial}. Depending on the nature of the reward-result gap, the reward model might need to be tailored to the agent's specific shortcomings~(e.g.\ be shaped away from unsafe states) rather than just purely capturing the human's intentions.

\section{Approaches}
\label{sec:approaches}

This section discusses a number of approaches that collectively may help to mitigate the problems discussed in \autoref{sec:challenges}.
These approaches should be thought of as directions to explore; more research is needed to figure out whether they are fruitful.

\subsection{Online feedback}
\label{ssec:online-feedback}

Preliminary experiments show failure modes when the reward model is not trained \emph{online}, i.e.\ in parallel with the agent~\citep{christiano2017deep,ibarz2018}. In these cases the agent learns to exploit reward models that are frozen. Because there is no additional user feedback, loopholes in the reward model that the agent discovers cannot be corrected.

If we provide the agent with reward feedback online, we get a tighter feedback loop between the user's feedback and the agent's behavior. This allows the reward model to be adapted to the state distribution the agent is visiting, mitigating some distributional shift problems. Moreover, with online feedback the user can spot attempts to hack the reward model and correct them accordingly.
Ideally, we would like the agent to share some responsibility for determining when feedback is needed, for instance based on uncertainty estimates (\autoref{ssec:uncertainty-estimates}), since otherwise providing relevant feedback in a timely manner could be prohibitively expensive.

\subsection{Off-policy feedback}
\label{ssec:off-policy-feedback}

When training the agent with feedback on its behavior, this feedback is only reactive, based on outcomes that have already occurred. To prevent unacceptable outcomes and reward hacking, we need to be able to communicate that certain outcomes are undesirable \emph{before they occur}. This requires the reward model to be accurate \emph{off-policy}, i.e.\ on states the agent has never visited~\citep{everitt2017reinforcement}. If off-policy feedback is used in conjunction with model-based RL~(\autoref{ssec:model-based-RL}), the agent can successfully avoid unsafe behavior that has never occurred.

The user could proactively provide off-policy feedback in anticipation of potential pitfalls~\citep{abel2017agent}.
Off-policy feedback could be elicited by using a generative model of the environment to create hypothetical scenarios of counterfactual events. However, generative modelling of states the agent has never visited might be very difficult because of the incurred distributional shift; the resulting videos might miss important details or be incomprehensible to humans altogether. Therefore it might be more feasible to provide off-policy feedback on an abstract level, for example using natural language~\citep{yeh2018bridging}. This is analogous to how humans can learn about bad outcomes through story-telling and imagination~\citep{riedl2016using}.

\subsection{Leveraging existing data}
\label{ssec:existing-data}

A large volume of human-created video data and prose is already readily available. Most of this data currently does not have high-quality text annotations and thus cannot be directly used as reward labels. Nevertheless, it contains a lot of useful information about human intentions~\citep{riedl2016using}. There are at least two approaches to leverage this existing data: using unsupervised learning~(such as unsupervised pretraining or third-person imitation learning; \citealp{stadie2017third}) or by manually annotating it.\footnote{For example, the total length of all movies on the Internet Movie Database longer than 40min is about 500,000 hours~\citep{peter2014watching}. Assuming a 10x overhead and \$10 per hour, this data would cost ca.\ \$50~million to annotate.}

\subsection{Hierarchical feedback}
\label{ssec:hierarchical-feedback}

The same arguments that support hierarchical RL~\citep{dayan1993feudal,sutton1999between,vezhnevets2017feudal} also encourage having a hierarchical decomposition of the reward model. This would allow the user to provide both low-level and high-level feedback. Both hierarchical RL and hierarchical reward models should be quite natural to combine: if the temporal hierarchies between agent and reward model align, then at each level of the hierarchy the reward model can train the corresponding level of the agent. This might help bypass some very difficult long-term credit assignment problems.

For example, recall the fantasy author task from \autoref{ssec:recursive-reward-modeling}. The low-level feedback would include spelling, fluency, and tone of language while high-level feedback could target plot and character development that cannot be provided on a paragraph level.

\subsection{Natural language}
\label{ssec:natural-language}

Since we want agents to be able to pursue and achieve a wide variety of goals in the same environment and be able to specify them in a way that is natural to humans, we could model the reward function as conditioned on natural language instructions~\citep{bahdanau2018learning}. These natural language instructions can be viewed as human-readable task labels. Moreover, they provide a separate privileged channel that should be easier to protect and harder to spoof than any instructions that are received through the observation channel.

In addition to providing task labels, we could also make natural language a more central part of the agent's architecture and training procedure. This has a number of advantages.
\begin{enumerate}
\item Natural language is a natural form of feedback for humans. If we can learn to translate natural language utterances into the rigid format required for the data set the reward model is trained on, this would allow users to give feedback much more efficiently.
\item Natural language has the potential to achieve better generalization if the latent space is represented using language~\citep{andreas2018learning} and possibly generalize in a way that is more predictable to humans. This might also help to mitigate distributional problems for the reward model~(\autoref{ssec:training-distribution}): if the training distribution is reasonably dense in the space of natural language paragraphs, this might make out-of-distribution inputs very rare.
\item Natural language might lead to substantially better interpretability. Especially for abstract high-level concepts, natural language might be much better suited than visual interpretability techniques~\citep{olah2018the}. However, by default the reward model's representations might not correspond neatly with short natural language expressions and will probably need to be trained particularly for this target~(without producing rationalizations).
\end{enumerate}

\subsection{Model-based RL}
\label{ssec:model-based-RL}

A \emph{model-based} RL agent learns an explicit model of the environment which it can use with a planning algorithm such as Monte Carlo tree search~\citep{abramson1987expected,kocsis2006bandit}. If we are training a model-based agent, the reward model can be part of the search process at planning time. This allows the agent to use \emph{off-policy} reward estimates, estimated for actions it never actually takes, provided that the reward model is accurate off-policy~(\autoref{ssec:off-policy-feedback}). This has a number of advantages:
\begin{enumerate}
\item The agent can avoid unacceptable outcomes~(\autoref{ssec:unacceptable-outcomes}) by discovering them during planning.
\item The agent's model could be used to solicit feedback from the user for outcomes that have not yet occured.
\item The agent can adapt to changes in the reward model more quickly because it can backup these changes to value estimates using the model without interaction with the environment.
\item Model-based approaches enable principled solutions to the reward tampering problem~(\autoref{ssec:reward-hacking}) by evaluating future outcomes with the current reward model during planning~\citep[Part~II]{everitt2018}. Agents that plan this way have no incentive to change their reward functions~\citep{schmidhuber2007godel,omohundro2008basic}, nor manipulate the register holding the reward signal~\citep[Sec.~6.3]{everitt2018}.
\end{enumerate}

\subsection{Side-constraints}
\label{ssec:side-constraints}

In addition to learning a reward function, we could also learn side-constraints for low-level or high-level actions~(\emph{options}; \citealp{sutton1999between}) to prevent unacceptable outcomes. Blocking actions can be more effective than discouraging them with large negative reward since negative rewards could be compensated by larger rewards later~(such as in the case of reward hacking). This problem could be amplified by errors in the agent's model of the world.

The same techniques described here for training a reward model should apply to train a model that estimates side-constraints and blocks low-level actions~\citep{saunders2018trial} or enforces constraints during policy updates~\citep{achiam2017constrained}. The main downside of this technique is that it puts additional burden on the human because they have to understand which actions can lead to unacceptable outcomes. Depending on the domain, this might require the human to be assisted by other agents. These agents could in turn be trained using recursive reward modeling~(\autoref{ssec:recursive-reward-modeling}).

\subsection{Adversarial training}
\label{ssec:adversarial-training}

To mitigate the effect of adversarially crafted inputs to neural networks~\citep{szegedy2013intriguing}, so far the empirically most effective strategy has been \emph{adversarial training}: training the model explicitly on adversarially perturbed inputs~\citep{madry2017towards,uesato2018adversarial,athalye2018obfuscated}.

However, it is unclear how to define adversarial perturbation rigorously in a general way~\citep{brown2018unrestricted, gilmer2018motivating}. To cover more general cases, we could train agents to explicitly discover weaknesses in the reward model and opportunities for reward hacking as well as the minimal perturbation that leads to an unacceptable outcome~\citep{uesato2018rigorous}. This is analogous to \emph{red teams}, teams whose objective is to find attack strategies~(e.g.\ security vulnerabilities) that an adversary might use~\citep{mulvaney2012red}.

The discovered failure cases can then be reviewed by the user and added to the feedback dataset. This might mean higher data requirements; so even if adversarial training fixes the problem, it might push the data requirements beyond affordable limits.

\subsection{Uncertainty estimates}
\label{ssec:uncertainty-estimates}

Another desirable feature of the reward model is an appropriate expression of uncertainty regarding its outputs.
Improving uncertainty estimates brings two benefits:
\begin{enumerate}
\item During training, it can help automate the process of soliciting feedback about the most informative states~\citep{krueger2016active, schulze2018active} using active learning~\citep{settles2012active}.
\item The agent can defer to the human or fall back to risk-averse decision making when uncertainty is large, for instance on inputs that do not resemble the training distribution~\citep{hendrycks2016baseline}.
\end{enumerate}

A number of recent works develop scaleable approximate Bayesian methods for neural networks, beginning with \citet{graves2011practical}, \citet{blundell2015weight}, \citet{kingma2015variational}, \citet{hernandez2015probabilistic}, and \citet{gal2016dropout}. So far model ensembles provide a very strong baseline~\citep{lakshminarayanan2017simple}.
Bayesian methods untangle irreducible uncertainty from `epistemic' uncertainty about which parameters are correct, which decreases with the amount of data~\citep{kendall2017uncertainties}; this distinction can help with active learning~\citep{gal2017deep}.

Other works aim to calibrate the predictions of neural networks~\citep{guo2017calibration}, so that their subjective uncertainty corresponds with their empirical frequency of mistakes.
While Bayesian methods can help with calibration~\citep{gal2017concrete}, they are insufficient in practice for deep neural networks~\citep{kuleshov2018accurate}.
Well-calibrated models could engage risk-averse decision making, but handling out-of-distribution states reliably would require higher quality uncertainty estimates than current deep learning techniques can provide~\citep{shafaei2018does}.

\subsection{Inductive bias}
\label{ssec:inductive-bias}

Finally, a crucial aspect of reward modeling is the inductive bias of the reward model.
Since we cannot train the reward model and the agent on all possible outcomes, we need it to generalize appropriately from the given data~\citep{zhang2017understanding,zhang2018study}.
The success of deep learning has been attributed to inductive biases such as distributed representations and compositionality, which may also be necessary in order to defeat the `curse of dimensionality'~\citep{bengio2013representation}.
Yet further inductive biases are necessary to solve many tasks; for instance, convolutional neural networks~\citep{lecun1990handwritten} vastly outperform multilayer perceptrons in computer vision applications because of their spatial invariance.

Solving reward modeling may require non-standard inductive biases; for instance modern deep networks typically use piece-wise linear activation functions~\citep{nair2010rectified, glorot2011deep, goodfellow2013maxout, xu2015empirical}, which generalize linearly far from training data~\citep{goodfellow2015}, meaning estimated reward would go to positive or negative infinity for extreme inputs.
The inductive bias of deep models can be influenced by the architecture, activation functions, and training procedure.
A growing body of work targets \emph{systematic generalization} in deep models. Examples include
modularity~\citep{anonymous2019systematic},
recursion~\citep{cai2017making},
graph structure~\citep{battaglia2018relational} or natural language~\citep{andreas2018learning} in the latent space,
differentiable external memory~\citep{graves2016hybrid},
or neural units designed to perform arbitrary arithmetic operations~\citep{trask2018neural}.

\section{Establishing trust}
\label{sec:establishing-trust}

Suppose our research direction is successful and we figure out how to train agents to behave in accordance with user intentions. How can we be confident that the agent we are training is indeed sufficiently aligned? In other words, how can we be confident that we have overcome the challenges from \autoref{sec:challenges} and that the agent's behavior sufficiently captures human intentions? This requires additional techniques that allow us to gain \emph{trust} in the agents we are training.

An ambitious goal is to enable the production of \emph{safety certificates}, artifacts that serve as evidence to convince a third party to trust our system. These safety certificates could be used to prove responsible technology development, defuse competition, and demonstrate compliance with regulations. A safety certificate could take the form of a score on a secret test suite held by a third party, evidence of interpretability properties, or a machine-checkable formal proof of correctness with respect to some established specification, among others.
A few general approaches for building trust in our models are discussed below.

\begin{figure}[t]
\centering
\includegraphics[width=0.8\textwidth]{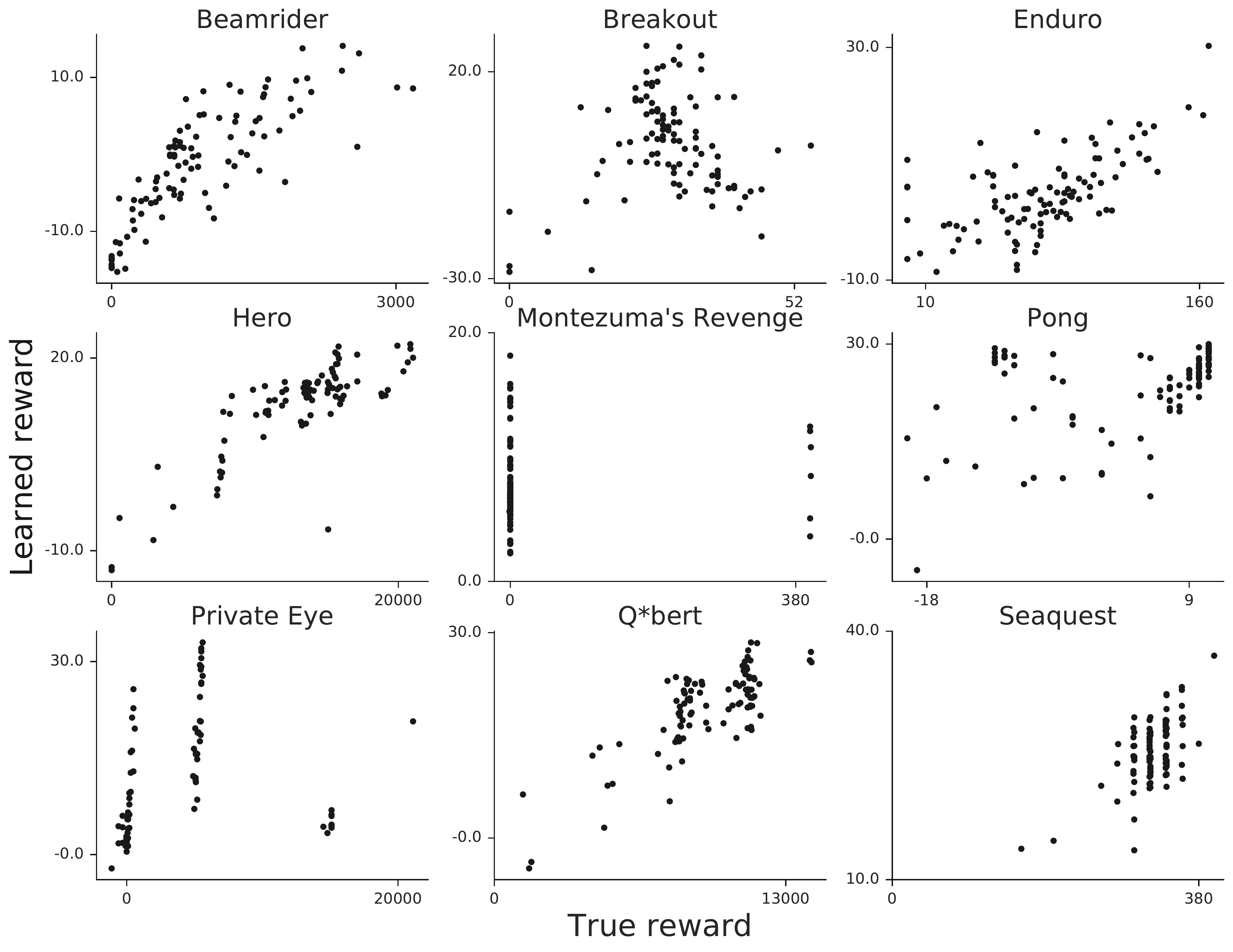}
\caption{Alignment of learned reward functions in 9 Atari games: Scatterplot showing the correlation of the reward learned from user preferences~(y-axis) with the true Atari reward~(x-axis) averaged over 1000~timesteps. For a fully aligned reward function, all points would be on a straight line.
In these experiments the reward model is well-aligned in some games like Beamrider, Hero, and Q*bert, and poorly aligned in others like Private Eye, Breakout, and Mondezuma's Revenge.
Reproduced from \citet{ibarz2018}.
}
\label{fig:alignment-atari}
\end{figure}

\paragraph{Design choices.}
Separating learning the objective from learning the behavior allows us to achieve higher confidence in the resulting behavior because we can split trust in the reward model from trust in the policy. For example, we can measure how well the reward function aligns with the task objective by evaluating it on the user's feedback~(see \autoref{fig:alignment-atari}). If we understand and trust the reward model, we know what the agent is `trying' to accomplish. If \autoref{ass:evaluation} is true, then the reward model should be easier to interpret and debug than the policy.

Another design choice that could increase trust in the system is to split our policy into two parts: a \emph{plan generator} and a \emph{plan executor}. The plan generator produces a human-readable plan of the current course of action. This plan could be very high-level like a business plan or a research proposal, or fairly low-level like a cooking recipe. This plan can then optionally be reviewed and signed off by the user. The plan executor then takes the plan and implements it.

Clean, well-understood design choices on training setup, model architecture, loss function, and so on can lead to more predictable behavior and thus increase our overall confidence in the resulting system~(as opposed to e.g.\ training a big blob of parameters end-to-end).
Especially if we manage to formally specify certain safety properties~\citep{orseau2016safely,krakovna2018measuring}, we can then make them an explicit part of our agent design.

\paragraph{Testing.}
Evaluation on a separate held-out test set is already common practice in machine learning. For supervised learning, the performance of a trained model is estimated by the empirical risk on a held-out test set which is drawn from the same data distribution. This practise can readily be applied to reward model~\citep{ibarz2018} and policy, e.g.\ on a set of specifically designed simulated environments~\citep{leike2017ai} or even adversarially where an attacker explicitly tries to cause misbehavior in the agent~\citep{uesato2018rigorous}.

\paragraph{Interpretability.}
Interpretability has been defined as the ability to explain or to present in understandable terms to a human~\citep{doshi2017towards}. Currently widely used deep neural networks are mostly black boxes, and understanding their internal functionality is considered very difficult. Nevertheless, recent progress provides reason for optimism that we will be able to make these black boxes increasingly transparent. This includes preliminary work on visualizing the latent state space of agents using t-SNE plots~\citep{zahavy2016graying,jaderberg2018human}, examining what agents attend to when they make decisions~\citep{greydanus2018visualizing},
evaluating models' sensitivity to the presence/intensity of high-level human concepts~\citep{kim2017tcav},
optimizing a model to be more interpretable with humans in the loop~\citep{lage2018human}, translating neural activations into natural language on tasks also performed by humans~\citep{andreas2017translating}, and combining different interactive visualization techniques~\citep{olah2018the}, to name only a few.

\paragraph{Formal verification.}
Recent progress on model checking for neural networks opens the door for formal verification of trained models~\citep{katz2017reluplex}. The size of verified models has been pushed beyond MNIST-size to over a million parameters~\citep{dvijotham2018dual,wong2018scaling}, which indicates that verifying practically sized RL models might soon be within reach.
If formal verification can be scaled, we could attempt to verify properties of policies~\citep{bastani2018verifiable} and reward functions with respect to a high-level specification, including off-switches, side-effects, and others mentioned in \autoref{sec:reward-modeling}. If \autoref{ass:learn-safety} from \autoref{sec:introduction} is true, then this specification does not have to be manually written, but instead can be provided by a separately learned model. However, in this case a formal correctness proof is only as useful as this learned specification is accurate.

To make the verification task easier, our models could be trained to be more easily verifiable~\citep{dvijotham2018training}. However, this opens the door for degenerate solutions that exploit loopholes in the learned specification. This is analogous to problems with reward hacking~(\autoref{ssec:reward-hacking}) which train a policy to optimize a frozen reward model~(\autoref{fig:reward-hacking}). Circumventing this problem could be done using the same techniques that have been successful for reward hacking, such as learning the specification online using user feedback~(\autoref{ssec:online-feedback}).

\paragraph{Theoretical guarantees.}
Finally, even more ambitious would be the development of theoretically well-founded scalable learning algorithms that come with \emph{probably approximately correct}~\citep{dziugaite2017computing} or \emph{sample complexity} guarantees, capacity statements, well-calibrated uncertainty estimates, etc.~\citep{veness2017online}. Unfortunately, currently there is a dire lack of any such guarantees for the popular deep neural network architectures and training techniques.

\section{Alternatives for agent alignment}
\label{sec:alternatives}

The research direction we outline in this paper is not the only possible path to solve the agent alignment problem. While we believe it is currently the most promising one to explore, it is not guaranteed to succeed. Fortunately there are a number of other promising directions for agent alignment. These can be pursued in parallel or even combined with each other. This section provides an overview and explains how our approach relates to them. Our list is not exhaustive; more directions are likely to be proposed in the future.

\subsection{Imitation learning}
\label{ssec:imitation-learning}

One strategy to train aligned agents could be from imitating human behavior~\citep{pomerleau1991efficient,abbeel2004apprenticeship,ho2016generative,finn2016guided}.
An agent imitating aligned human behavior sufficiently well should be aligned as well. The following caveats apply:

\begin{itemize}
\item \emph{Amount of data}. While feedback can often be provided by non-experts, the data for human imitation has to be provided by experts on the task. This might be much more expensive data and it is not clear if we need more or less than for reward modeling.
\item \emph{Cognitive imitation.} It is possible that a lot of cognitively demanding tasks that humans do rely on very high-level intuition, planning, and other cognitive processes that are poorly reflected in human actions. For example, a crucial insight for solving a problem might be gained from drawing an analogy with a different problem encountered in a different domain. This might be hard to replicate and predict from data about human actions alone.
\item \emph{Generalization.} In order to be useful, our agent trained with imitation learning needs to showcase persistently high-quality behavior, even in the face of novel situations. Analogous to \autoref{ass:evaluation}, generalizing learned reward functions might be easier than generalizing behavior~\citep{bahdanau2018learning}.
\item \emph{Performance.} It is generally difficult to outperform humans using imitation learning alone~\citep{hester2018deep}: even a perfect imitator can only perform as well as the source it is imitating; superhuman performance typically comes from executing human action sequences faster and more reliably by smoothing out inconsistencies in human behavior~\citep{aytar2018playing}.
\end{itemize}

Therefore imitation learning is unlikely to be competitive with other strategies to train agents in the longer term. However, it might be sufficient to act as a `stepping stone': agents trained with imitation learning might act as `research assistants' and help scale up other alignment efforts. Therefore it should be considered as a strong alternative to our research strategy.

\subsection{Inverse reinforcement learning}
\label{ssec:inverse-rl}

We can view a reinforcement learning algorithm as a mapping from a reward function to behavior. The inverse of that mapping takes agent behavior as input and produces a reward function; this is known as \emph{inverse reinforcement learning}~(IRL; \citealp{russell1998learning,ng2000algorithms}). In this sense, inverse reinforcement learning can be viewed as one approach to reward modeling that takes feedback in the form of trajectories of behavior. However, taken as it is, it had two shortcomings:

\begin{enumerate}
\item IRL is an under-constrained problem because the reward function is not uniquely identifiable~(not even up to affine-linear transformation) from behavior alone~\citep{ng2000algorithms}; for example, $R = 0$ is always a solution. If we assume the human is fully rational and the agent can design a sequence of tasks for the human, then the reward function can be identified~\citep{amin2017repeated}. Even some assumptions about the human's rationality can be relaxed~\citep{evans2016learning}, but in full generality the inverse reinforcement learning problem becomes impossible to solve~\citep{armstrong2017impossibility}.
\item It assumes the human is acting to optimize their reward directly, even when this is an inefficient way of communicating their preferences. For instance, it is much easier for a human to state `I would like you to make me coffee every morning at 8am' than it is for the human to make themselves coffee at 8am several days in a row.
\end{enumerate}

\subsection{Cooperative inverse reinforcement learning}
\label{ssec:cirl}

Motivated by this second shortcoming of IRL, \citet{hadfield2016cooperative} propose \emph{cooperative inverse reinforcement learning}~(CIRL).
CIRL is a formal model of reward modeling as a two player game between a user and an agent which proceeds as follows.
\begin{enumerate}
\item The user and the agent begin with a shared prior over the user's reward function,
\item the user then observes their reward function, and finally
\item both user and agent execute policies to optimize the user's reward function.
\end{enumerate}
An optimal solution to a CIRL game would use the common knowledge of the user and the agent to compute a policy for the agent (to be executed in step 3), and a mapping from reward function to policy for the user. Then upon observing their reward function in step 2, the user should select the corresponding policy for them to execute in step 3.
Both the user and the agent have to choose behaviors which trade off between
\begin{enumerate*}[label={(\arabic*)}]
\item communicating the user's reward function to the agent and
\item directly maximizing the user's expected reward.
\end{enumerate*}

We make two observations about CIRL as an approach to agent alignment that highlight that CIRL abstracts away from some important details.
First, the performance of a CIRL algorithm will depend on the quality of the prior over reward functions. In essence, CIRL replaces the problem of specifying a reward function with specifying a prior over reward functions.
Second, computing the optimal solution to the CIRL problem is not realistic, since we cannot prescribe exactly how the user should interact with the agent. In other words, an efficient solution to a CIRL game might employ a strategy that transmits the parameters from the user to the agent, followed by a normal RL algorithm executed by both the user and the agent~(since the reward is now fully observable to both). But if the user were able to observe their reward function, they could just specify this to an RL agent directly. In other words, one of the difficulties of agent alignment is that the reward function is not directly available to the user in the first place:  users are usually not very aware of all of their preferences, and it might instead be easier for them to communicate through revealed preferences~\citep{samuelson1938note}.

Nevertheless, CIRL incorporates two important insights into the alignment problem that also motivate our research direction:
\begin{enumerate}
\item Constructing agents to optimize a \emph{latent} reward function can help align them on tasks where we cannot consistently provide reward feedback about all state-action pairs as the agent is visiting them.
\item A key challenge of the agent alignment problem is finding efficient ways to communicate the user's intentions to learning agents.
\end{enumerate}

\subsection{Myopic reinforcement learning}
\label{ssec:myopic-rl}

Myopic RL agents only maximize reward in the present timestep instead of a (discounted) sum of future rewards.  This means that they are more short-sighted and thus not incentivized to execute long-term plans or take actions that are bad in the short-term in order to get a long-term benefit. In particular, myopic RL agents might be less prone to some of the design specification problems mentioned in \autoref{sec:reward-modeling}, since causing them might take several time-steps to pay off for the agent.

There are two main myopic RL algorithms. TAMER~\citep{Knox09,knox2012learning,Warnell17} is a collection of algorithms that learn a policy from human value feedback, i.e.\ take actions that maximize expected feedback in the next step~(possibly with short temporal smoothing). COACH~\citep{macglashan2017interactive,arumugam2018deep} is an algorithm that trains a policy from feedback in the form of an \emph{advantage function}~\citep{sutton1998}.

In contrast to imitation learning, the user does not have to be able to produce the desired behavior, just be able to reward the individual actions that bring it about. For example, using TAMER or COACH, a user could teach an agent to perform a backflip without being able to do one themself.
However, while myopic RL may increase alignment, is also comes with performance drawbacks. Training myopic RL agents puts the burden of solving the credit assignment problem onto the user, limiting the agent's potential for ingenuity and thus performance, and also leaving the user responsible for avoiding long-term negative consequences.

Despite these limits, myopic RL agents might be sufficient for some tasks where credit assignment is reasonably easy for humans. They might also be used as building blocks in more capable training regimes, for instance in iterated amplification~\citep{christiano2018supervising}.

\subsection{Imitating expert reasoning}
\label{ssec:amplification}

Another alternative is to train a model to imitate expert reasoning.
The imitation can happen at a level of granularity decided by the expert and could include `internal' reasoning steps that the expert would not typically perform explicitly. This expert reasoning can then be improved and accelerated~\citep{christiano2018supervising,evans2018predicting,stuhlmueller2018factored}.

The basic idea is best illustrated with a question answering system. The input to the system is a question $Q$ and its output an answer $A$. For simplicity we can treat both $Q$ and $A$ as natural language strings. The system can call itself recursively by asking subquestions $Q_1, \ldots, Q_k$, receiving their answers $A_1, \ldots, A_k$, and composing them into the answer $A$.

For example, consider the question $Q$ `How many pineapples are there in Denmark?' To give an approximate answer, we could make a \emph{Fermi estimate} by asking the subquestions `What is the population of Denmark?', `How many pineapples does the average Dane consume per year?', and `How long are pineapples stored?' These subquestions are then answered recursively and their answers can be composed into an answer to the original question $Q$.

We could train a model to answer questions $Q$ recursively by using the same reasoning procedure as the expert using imitation learning~(\autoref{ssec:imitation-learning}). This model can then be improved using a variety of methods:

\begin{itemize}
\item Running many copies of this model in parallel and/or at greater speed.
\item Training a new model to predict answers to questions without having to expand the subquestions, akin to using a value network to the estimate the result of a tree search~\citep{anthony2017thinking,silver2017mastering}.
\item Making the expert reasoning more coherent under reflection. For example, by searching for inconsistencies in the expert's reasoning and resolving them.
\end{itemize}

If we believe expert reasoning is aligned with the user, then we could hope that the resulting improved model is also aligned.
This training procedure aims to achieve better interpretability and greater trust in the resulting agents than recursive reward modeling~(\autoref{ssec:recursive-reward-modeling}). However, learning expert reasoning might not be economically competitive with recursive reward modeling, depending on how good the expert's reasoning is and whether \autoref{ass:evaluation} holds for the task at hand.

Even though both are an instance of the more general framework of iterated amplification~\citep{christiano2018supervising}, recursive reward modeling as described in \autoref{ssec:recursive-reward-modeling} does not try to model expert reasoning explicitly. Instead, recursive reward modeling only requires users to evaluate outcomes. Nevertheless, it relies on decomposition of the evaluation task which has similarities to the decompositional reasoning described here.
When using recursive reward modeling users have the \emph{option} to provide feedback on the cognitive process that produced outcomes, but they are not required to do so.
Moreover, this feedback might be difficult to provide in practice if the policy model is not very interpretable.

\subsection{Debate}
\label{ssec:debate}

\citet{irving2018ai} describe an idea for agent alignment that involves a two-player zero-sum game in which both players are debating a question for the user. The two players take turns to output a short statement up to a turn limit. At the end of the game the user reads the conversation transcript and declares the player who contributed the most true and useful statements the winner.

The debate proposal involves training an agent with self play~\citep{silver2016mastering} on this debate game. In order to become aligned, this agent needs to be trained in a way that it converges to a Nash equilibrium in which both instances of the agent try to be helpful to the user. The central assumption of debate is that it is easier for the agent to tell the truth than it is to lie. If this assumption holds, then the dynamics of the game should incentivize the agent to provide true and useful statements.

The authors provide initial experiments on the MNIST dataset in which the debating agents manage to boost the accuracy of a sparse classifier that only has access to a few of the image's pixels. While these initial experiments are promising, more research is needed in order to determine whether debate is a scalable alignment approach. We need more empirical evidence to clarify, among others, the following two questions.
\begin{enumerate}
\item Does the central assumption of debate hold outside domains of easily fact-checkable statements?
\item Can the humans accurately judge the debate even if the debaters have superior persuasion and deception ability?
\end{enumerate}

\subsection{Other related work}
\label{ssec:other-related-work}

Many of the practical challenges to reward modeling we raise here have already been discussed by \citet{amodei2016concrete}: safe exploration, distributional shift, side-effects, and reward hacking. In particular, the authors highlight what they call the scalable oversight problem, how to train an RL agent with sparse human feedback. This can be understood as a more narrow version of the alignment problem we are aiming to solve here.
In a similar spirit, \citet{taylor2016alignment} survey a number of high-level open research questions on agent alignment. Most closely related to our approach are what the authors call informed oversight~(building systems that help explain outcomes), generalizable environmental goals~(defining objective functions in terms of environment states), and averting instrumental incentives~(preventing the system from optimizing for certain undesirable subgoals).

\citet{soares2017agent} outline a research agenda of a very different flavor. Their research problems are quite paradigm-agnostic and instead concern the theoretical foundations of mathematical agent models. In particular, many of their problems aim to address perceived difficulties in applying current notions of optimal behavior to agents which are part of their environment~\citep{orseau2012space} and thus may not remain cleanly delineated from it~\citep{demski2018embedded}.
The authors seek the formal tools to ask questions about or relevant to alignment in theory, such as when provided with a halting oracle~\citep{hutter2005}. These formal tools could be necessary for formal verification of agents designing upgraded versions of themselves. Yet while there has been some of progress on this research agenda~\citep{barasz2014robust,leike2016formal,garrabrant2016logical,everitt2018}, some questions turned out to be quite difficult. But even if we had formal solutions to the problems put forth by \citeauthor{soares2017agent}, there would still persist a gap to transfer these solutions to align agents in practice. For now, answers to these research questions should be understood more as intuition pumps for practical alignment questions rather than direct solutions themselves~\citep{garrabrant2018}.

See \citet{everitt2018agi} for more in-depth survey and literature review.

\section{Discussion}
\label{sec:discussion}

\paragraph{Summary.}
The version of the agent alignment problem we are aiming to solve involves aligning a single agent to a single user~(\autoref{sec:alignment-problem}).
Instead of attempting to learn the entire preference payload, we outline an approach for enabling the user to communicate their intentions to the agent for the task at hand so that it allows them to trust the trained agent.

Our research direction for agent alignment is based on scaling reward modeling~(\autoref{sec:roadmap}). This direction fits well into existing efforts in machine learning because it can benefit from advances in the state of the art in supervised learning~(for the reward model) and reinforcement learning~(for the policy). Building on previous work~(\autoref{sec:alternatives}), we provide significantly more detail, including the main challenges~(\autoref{sec:challenges}) and concrete approaches to mitigate these challenges~(\autoref{sec:approaches}) and to establish trust in the agents we train~(\autoref{sec:establishing-trust}). In essence, this document combines existing efforts on AI safety problems by providing one coherent narrative around how solving these problems could enable us to train aligned agents beyond human-level performance.

\paragraph{Concrete research projects.}
Our research direction is `shovel-ready' for empirical research today. We can set up experiments with deep reinforcement learning agents: getting empirical data on the severity of the challenges from \autoref{sec:challenges};
prototyping solution ideas from \autoref{sec:approaches};
scaling reward modeling to more difficult tasks;
pushing the frontiers on (adversarial)~testing, interpretability, formal verification, and the theory of deep RL. Moreover, we can readily use any existing RL benchmarks such as games or simulated environments that come with pre-programmed reward functions: By hiding this reward function from the algorithm we can pretend it is unavailable, but still use it for synthetically generated user feedback~\citep{christiano2017deep} as well as the evaluation of the learned reward model~\citep{ibarz2018}.

\paragraph{Outlook.}
There is enormous potential for ML to have a positive impact on the real world and improve human lives. Since most real-world problems can be cast in the RL framework, deep RL is a particularly promising technique for solving real-world problems. However, in order to unlock its potential, we need to train agents in the absence of well-specified reward functions. Just as proactive research into robustness of computer vision systems is essential for addressing vulnerabilities to adversarial inputs, so could alignment research be key to getting ahead of future bottlenecks to the deployment of ML systems in complex real-world domains. For now, agent alignment research is still in its early stages, but we believe that there is substantial reason for optimism. While we expect to face challenges when scaling reward modeling, these challenges are concrete technical problems that we can make progress on with targeted research.

\subsubsection*{Acknowledgments}

This paper has benefited greatly from discussions with many people at DeepMind, OpenAI, and the Future of Humanity Institute. For detailed feedback we are particularly grateful to Paul Christiano, Andreas Stuhlmüller, Ramana Kumar, Laurent Orseau, Edward Grefenstette, Klaus Greff, Shahar Avin, Tegan Maharaj, Victoria Krakovna, Geoffrey Irving, Owain Evans, Andrew Trask, Iason Gabriel, Elizabeth Barnes, Miles Brundage, Alex Zhu, Vlad Firoiu, Serkan Cabi, Richard Ngo, Jonathan Uesato, Tim Genewein, Nick Bostrom, Dario Amodei, Felix Hill, Tom McGrath, Borja Ibarz, Reimar Leike, Pushmeet Kohli, Greg Wayne, Timothy Lillicrap, Chad Burns, Teddy Collins, Adam Cain, Jelena Luketina, Eric Drexler, Toby Ord, Zac Kenton, and Pedro Ortega.

\footnotesize
\bibliographystyle{iclr2019_conference}
\bibliography{references}

\end{document}